\documentclass{llncs}
\usepackage{llncsdoc}

\usepackage[T1]{fontenc}
\usepackage[utf8]{inputenc}
\usepackage{authblk}
\usepackage{makeidx} 
\usepackage{booktabs}
\usepackage{amsmath}
\usepackage{amsfonts}
\usepackage{amssymb}
\usepackage{mathtools}
\usepackage{times}
\usepackage[square,sort,comma,numbers]{natbib}
\usepackage{algorithm}
\usepackage{algorithmic}
\usepackage{graphicx} 
\usepackage{subfigure} 
\graphicspath{{figures/}} 
\usepackage{epstopdf}
\usepackage{sidecap}
\sidecaptionvpos{figure}{c}
\usepackage{comment}

\usepackage[usenames]{color}

\usepackage{enumitem}
\usepackage{bbold}
\usepackage{multirow}
\usepackage{pbox}
\usepackage{array}
\newcolumntype{L}[1]{>{\raggedright\let\newline\\\arraybackslash\hspace{0pt}}m{#1}}
\newcolumntype{C}[1]{>{\centering\let\newline\\\arraybackslash\hspace{0pt}}m{#1}}
\newcolumntype{R}[1]{>{\raggedleft\let\newline\\\arraybackslash\hspace{0pt}}m{#1}}

\AtBeginDocument{}

\begin{document}
\frontmatter               
\pagestyle{headings}  

\title{Differentially Private Variational Dropout}
\titlerunning{Differentially Private Variational Dropout}  
%
\author{Beyza Ermis  \and A. Taylan Cemgil }
\authorrunning{Beyza Ermis et al.} 
\tocauthor{Beyza Ermis}
\institute{Bo\u{g}azi\c ci University, 34342, Bebek, \.Istanbul, Turkey,\\
\email{beyza.ermis@boun.edu.tr, taylan.cemgil@boun.edu.tr}
}

\maketitle              

\begin{abstract}
Deep neural networks with their large number of parameters are highly flexible learning systems. The high flexibility in such networks brings with some serious problems such as overfitting, and regularization is used to address this problem. A currently popular and effective regularization technique for controlling the overfitting is dropout. Often, large data collections required for neural networks contain sensitive information such as the medical histories of patients, and the privacy of the training data should be protected. In this paper, we modify the recently proposed variational dropout technique which provided an elegant Bayesian interpretation to dropout, and show that the intrinsic noise in the variational dropout can be exploited to obtain a degree of differential privacy. The iterative nature of training neural networks presents a challenge for privacy-preserving estimation since multiple iterations increase the amount of noise added. We overcome this by using a relaxed notion of differential privacy, called concentrated differential privacy, which provides tighter estimates on the overall privacy loss. We demonstrate the accuracy of our privacy-preserving variational dropout algorithm on benchmark datasets.

\keywords{Dropout, Variational Inference, Differential Privacy}
\end{abstract}

\section{Introduction}
\label{sec:intro}
Deep neural networks (DNN) have recently generated significant interest, largely due to their successes in several important learning applications, including image classification, language modeling and many more (e.g.,~\cite{MaddisonHSS14,NIPS2015_5635,HeZRS15}).
The success of neural networks is directly related to the availability of large and representative datasets for training. However, these datasets are often collected from people, such as their tastes and behavior as well as medical health records, and present obvious privacy issues. Their usage requires methods that provide precise privacy guarantees while meeting the demands of the applications.

Overfitting is another challenge in deep neural networks, since DNNs can model complex prediction functions using a large number of parameters.
It is often difficult to optimize these functions due to the potentially large number of local minimas in the space of parameters, and standard optimization techniques are prone to getting stuck in a local minimum which might be far from the global optimum. A popular regularization technique to avoid such local minima is dropout~\cite{Hinton2012,wang2013fast,Srivastava2014} which introduces noise into a model and optimizes loss function under stochastic setting. 
Recently, it was shown that variational dropout can be treated as a generalization of Gaussian dropout~\cite{kingma2015variational} and this method can be used to tune each weight's individual dropout rates~\cite{molchanov2017variational}. Besides, regularization techniques, specifically dropout, may hide details some of the training data. These features of the dropout method inspire us to use dropout in order to provide a theoretical guarantee for the privacy protection on neural networks.

We provide a general framework for privacy-preserving variational dropout algorithm by exploiting the inherent randomization of the dropout. 
Differential privacy (DP) is currently a widely accepted privacy definition~\cite{differential-privacy,dwork2010differential} and we use DP to provide a formalization to the privacy protection of the proposed algorithm. The main principle of DP is to ensure that an adversary should not be able to reliably infer whether or not a particular individual is participating in a database, even with unlimited computational power and access to every entry except for that particular individual's data. 
This can be accomplished through adding noise into an algorithm at different stages such as adding noise to data itself or changing the objective function to be optimized. 
In order to design efficient differentially private algorithms, one need to design a noise injection mechanism such that there is a good trade-off between privacy and utility. 
However, iterative algorithms such as variational inference accumulate the privacy loss at each access to the training data, and the number of iterations required to guarantee accurate posterior estimates causes high cumulative privacy loss.  
Therefore, our algorithm uses the zCDP composition analysis~\cite{bun2016concentrated} that is inspired by concentrated differential privacy (CDP)~\cite{dwork2016concentrated} which is a recently proposed notion of the differential privacy. CDP is well suited for iterative algorithms since it provides high probability bounds for cumulative privacy loss and requires adding much less noise for the same expected privacy guarantee compared to the DP.

In this paper, we study Variational dropout in the case when we tune individual dropout rates for each weight of neural network to provide measurable privacy guarantee.  
Our main contributions can be summarized as follows:
\begin{itemize}
\item We first extend the dropout algorithm to protect the privacy of the training data of DNNs. We use the intrinsic additive noise of the recently proposed variational dropout algorithm~\cite{molchanov2017variational}, then we explore and analyze that under what conditions dropout helps in training DNNs within a modest privacy budget.
\item In order to use the privacy budget more efficiently over many iterations, our approach uses the zCDP composition combined with the privacy amplification effect due to subsampling of data, which significantly decrease the amount of additive noise for the same expected privacy guarantee compared to the standard DP analysis. 
\item We empirically show that for general single hidden layer neural network models, dropout helps to regularize network and improves accuracy while providing $(\epsilon,\delta)$-DP and zCDP. As our experiments illustrate, variational dropout with zCDP outperforms both the standard DP and the state-of-the-art algorithms, especially when the privacy budget is low. 
\end{itemize}

The rest of the paper is organized as follows. In Section~\ref{sec:related}, we survey the related studies on differentially private deep neural networks and we provide an overview of the relevant ingredients in Section~\ref{sec:backgr}. We formalize our setup, present the differentially private variational dropout algorithm and provide a  theoretical guarantee for differential privacy on the presented model in Section~\ref{sec:methods}. Section~\ref{sec:exp_res} presents our experimental results on some real datasets. We conclude with some open directions in Section~\ref{sec:conc}.

\section{Related Work}
\label{sec:related}
Differential privacy has been actively studied in many machine learning and data mining problems. Dwork \emph{et al.}~\cite{dwork2010differential} cover much of the earlier theoretical work and Sarwate \emph{et al.}~\cite{sarwate2013signal} review differentially private signal processing and machine learning studies. Problems that have been studied in the literature range from recommender systems~\cite{mcsherry2009differentially}, classification~\cite{Prateek2013} and empirical risk minimization applications~\cite{bassily2014private} such as logistic regression~\cite{rajkumar2012differentially} and Support Vector Machines~\cite{song2013stochastic}.

There are a number of works that address deep learning under differential privacy. Recently, Shokri and Shmatikov~\cite{ShokriS15} designed a system that enables multiple parties to train a neural-network model without sharing their datasets. Their key contribution is the selective sharing of model parameters during training that is based on perturbing the gradients of the SGD. Phan \emph{et al.}~\cite{Phan0WD16} proposed a different approach towards differentially private deep learning that focuses on learning autoencoders. Privacy relies on perturbing the objective functions of these autoencoders. Most recently, Papernot \emph{et al.}~\cite{PapernotAEGT16} proposed a method where privacy-preserving models are learned locally from disjoint datasets, and then combined in a privacy-preserving fashion. Our work is most closely related to the work by Abadi \emph{et al.}~\cite{Abadi2016}. They developed the moments accountant method to accumulate the privacy cost that provides a tighter bound for privacy loss than previous composition methods. Then, by using the moments accountant they propose a differentially private SGD algorithm to train a neural network by perturbing the gradients in SGD.

Two distinct types of mechanisms have been proposed for differentially private variational inference: perturbing the sufficient statistics of an exponential family model~\cite{park2016variational} and perturbing the gradients in optimization of variational inference~\cite{jalko2016differentially}. The second mechanism~\cite{jalko2016differentially} uses the moments accountant to perturb the gradients. Our goal is to apply the concentrated DP~\cite{dwork2016concentrated,bun2016concentrated}, which is closely related to the moments accountant, to gradient perturbation mechanism. We exploit the intrinsic randomized noise of the variational dropout and derive similar bounds to the moments accountant to strengthen the privacy guarantee in neural networks.

\section{Background}
\label{sec:backgr}

\subsection{Differential Privacy}
\label{sec:dp}
A natural notion of privacy protection prevents inference about specific records by requiring a randomized query response mechanism that yields similar distributions on responses of similar datasets. Formally, for any two possible input datasets $\mathcal{D}$ and $\mathcal{D}^\prime$ with the edit distance or Hamming distance $d(\mathcal{D}, \mathcal{D}^\prime)$, and any subset of possible responses $R$, a randomized algorithm $\mathcal{A}$ satisfies ($\epsilon$, $\delta$) differential privacy if:
\begin{align}
P(\mathcal{A}(\mathcal{D}) \in R) \leq e^{\epsilon} P(\mathcal{A}(\mathcal{D}^\prime) \in R) + \delta 
\label{eqn:dp}
\end{align}
If $\mathcal{D}$ and $\mathcal{D}^\prime$ are the same except one data point, then $d(\mathcal{D}, \mathcal{D}^\prime)$ = 1. 
($\epsilon$, $\delta$)-differential privacy ensures that for all adjacent $\mathcal{D}$, $\mathcal{D}^\prime$, the absolute value of the privacy loss will be bounded by $\epsilon$ with probability at least $1-\delta$. 
Here, $\epsilon$ controls the maximum amount of information gain about an individual's data given the output of the algorithm. When the positive parameter $\epsilon$ is smaller, the mechanism provides stronger privacy guarantee~\citep{differential-privacy}.    

\subsubsection*{Concentrated Differential Privacy: } Concentrated differential privacy (CDP) is a recent variation of differential privacy which is proposed to make privacy-preserving iterative algorithms more practical than DP while still providing strong privacy guarantees.
The CDP framework treats the privacy loss of an outcome,
\begin{align}
L^{(o)}_{(\mathcal{A}(\mathcal{D})\parallel \mathcal{A}(\mathcal{D}^\prime))} = \log\frac{P(\mathcal{A}(\mathcal{D})=o)}{P(\mathcal{A}(\mathcal{D}^\prime)=o)}
\end{align}
as a random variable.
Two CDP methods are proposed in the literature. The first one is ($\mu$, $\tau$)-mCDP~\cite{dwork2016concentrated} where $\mu$ is the mean of this privacy loss and the second one is $\tau$-zCDP which is proposed by Bun and Steinke in~\cite{bun2016concentrated}. $\tau$-zCDP~\cite{bun2016concentrated} arises from a connection between the moment generating function of $L(o)$ and the R\'enyi divergence between the distributions of $\mathcal{A}_{(\mathcal{D})}$ and $\mathcal{A}_{(\mathcal{D}^\prime)}$. Most of the DP mechanisms and applications can be characterized in terms of zCDP, but not in terms of mCDP; so we use zCDP as a tool for analyzing composition under the ($\epsilon$, $\delta$)-DP privacy definition, for a fair comparison between CDP and DP analyses.

\subsection{Variational Inference}
\label{sec:vi}
Assume we are given data $\mathcal{D} = \{d_i\}_{i=1}^N$ where $d_i$ = ($x_i$, $y_i$), with input object/feature $x_i \in \mathcal{R}^D$ and output label $y_i \in \mathcal{Y}$, with $\mathcal{Y}$ being the output discrete label space. A model characterizes the relationship from $x$ to $y$ with parameters (or weights) $\theta$.
Our goal is to tune the parameters $\theta$ of a model $p(y | x,\theta)$ that predicts $y$ given $x$ and $\theta$.
Bayesian inference in such a model consists of updating some initial belief over parameters $\theta$ in the form of a prior distribution $p(\theta)$, after observing data $\mathcal{D}$, into an updated belief over these parameters in the form of the posterior distribution $p(\theta | \mathcal{D})$.
The posterior distribution of a set of $N$ items is $p(\theta | \mathcal{D}) \propto p(\theta) \ p(\mathcal{D} | \theta)$ where the corresponding data likelihood is $p(\mathcal{D} | \theta) = \prod_{i=1}^N p(d_i | \theta)$.

Computing the posterior distribution is often difficult in practice as it requires the computation of analytically intractable integrals, so we need to use approximation techniques. One of such techniques is Variational Inference~\cite{Jordan99anintroduction} that turns the inference problem into an optimization problem, which is often more easy to tackle and to monitor convergence. In this approach, true posterior $p(\theta | \mathcal{D})$ is approximated with a variational distribution $q_\phi(\theta)$ that has a simpler form than the posterior. The optimal value of variational parameters $\phi$ is obtained through minimizing the Kullback-Leibler (KL) divergence between $q_\phi(\theta)$ and $p(\theta | \mathcal{D})$.
This is also equivalent to maximizing the so-called evidence lower bound (ELBO). 
Given joint distribution $p(\mathcal{D}, \theta) = p(\mathcal{D} | \theta) \ p(\theta)$, ELBO of $q_\phi$ is given as follows:
\begin{align}
\mathcal{L}(q_\phi) &= \int q_\phi(\theta) \log\left(\frac{p(\mathcal{D | \theta})}{q_\phi(\theta)}\right) = - D_{KL} \big(q_\phi(\theta) \parallel p(\theta)\big) + \mathcal{L}_{\mathcal{D}}(q_\phi)  \label{eq:ELBO} \\
\mathcal{L}_{\mathcal{D}}(q_\phi) &=  \left\langle \log p(\mathcal{D} | \theta) \right\rangle_{q_\phi(\theta)} = \sum_{(x_i,y_i)\in\mathcal{D}} \left\langle \log p(y_i | x_i,\theta) \right\rangle_{q_\phi(\theta)}  \label{eq:expectedLL}
\end{align}
where $\left\langle\cdot\right\rangle_{q_\phi(\theta)}$ is expectation taken w.r.t $q_\phi(\theta)$ and $\mathcal{L}_{\mathcal{D}}(q_\phi)$ is the expected log-likelihood.

\subsubsection{Stochastic Variational Inference}
\label{sec:svi}
An efficient method for minibatch-based optimization of the variational lower bound is the Stochastic Variational Inference (SVI) introduced in~\cite{HoffmanBWP13}. 
The basic trick in SVI is to parameterize the random parameters $\theta \sim q_\phi(\theta)$ as a differentiable function $\theta = f(\phi, \varepsilon)$ where $\varepsilon \sim p(\varepsilon)$ is a random noise variable. This new parameterization allows us to obtain an unbiased differentiable minibatch-based Monte Carlo estimator of $\mathcal{L}_{\mathcal{D}}(q_\phi)$ and $\bigtriangledown_\phi \mathcal{L}_{\mathcal{D}}(q_\phi)$:
\begin{align}
\mathcal{L}_{\mathcal{D}}(q_\phi) & \simeq \mathcal{L}_{\mathcal{D}}^{SVI}(q_\phi) = \frac{N}{S} \sum_{i=1}^S \log p(y^i | x^i, \theta=f(\phi,\varepsilon))  \label{eq:estimator} \\
\bigtriangledown_\phi \mathcal{L}_{\mathcal{D}}(q_\phi) & \simeq \mathcal{L}_{\mathcal{D}}^{SVI}(q_\phi) = \frac{N}{S} \sum_{i=1}^S \bigtriangledown_\phi \log p(y^i | x^i, \theta=f(\phi,\varepsilon)) 
\label{eq:gradEstimator}
\end{align}
where $(x_i,y_i)_{i=1}^S$ is a minibatch of data with $S$ random datapoints $(x_i,y_i) \sim \mathcal{D}$. 
The theory of stochastic approximation tells us that the performance of stochastic gradient optimization crucially depends on the variance of the gradients~\cite{robbins1951}. We follow~\cite{kingma2015variational} and use the \emph{Local Reparameterization Trick} that reduces the variance of this gradient estimator. The idea is to sample separate weight matrices for each data-point inside mini-batch by moving the noise from weights to activations~\cite{wang2013fast,kingma2015variational}.

\subsection{Variational Dropout}
\label{sec:vd}
Dropout is one of the most popular regularization techniques for neural networks which injects multiplicative random noise to the input of each layer during the training procedure. The formalization of dropout is given as:
\begin{align}
B = (A \odot \xi) \theta \qquad \text{with} \qquad  \xi_{i,j} \sim p(\xi_{i,j})
\end{align}
where $A$ denotes the matrix of input features, $\theta$ is the weight matrix for the current layer, and $B$ denotes the matrix of activations. The $\odot$ symbol denotes the elementwise (Hadamard) product of the input matrix with a matrix of independent noise variables $\xi$. The previous publications~\cite{Hinton2012,wan2013regularization,Srivastava2014} show that the weight parameters $\theta$ are less likely to overfit to the training data by adding noise to the input during optimization.

At first, Hinton \emph{et al.}~\cite{hinton2012improving} proposed the Binary Dropout where the elements of $\xi$ are drawn from a Bernoulli distribution with parameter $1-p$, hence each element of the input matrix is put to zero with probability $p$ that is also known as \emph{dropout rate}. Afterwards, the same authors proposed the  Gaussian Dropout using continuous noise $\xi_{i,j} \sim \mathcal{N}(1,\alpha=\frac{p}{1-p})$ with same relative mean and variance works as well or better~\cite{Srivastava2014}. It is important to use continuous noise instead of discrete one, because adding Gaussian noise to the inputs corresponds to putting Gaussian noise on weights. 

Then, Kingma \emph{et al.}~\cite{kingma2015variational} proposed \emph{Variational Dropout} that generalizes Gaussian dropout~\cite{Srivastava2014} with continuous noise as a variational method. It allows to set individual dropout rates for each neuron or layer and to tune them with a simple gradient descent based method. 
The main idea of this procedure is to search for posterior approximation in a specific family of distributions: $q(\theta_{i,j} | \phi_{i,j}, \alpha)=\mathcal{N}(\phi_{i,j},\alpha\phi_{i,j}^2)$. That is, putting multiplicative Gaussian noise $\xi_{i,j} \sim \mathcal{N}(1,\alpha)$ on weight $\theta_{i,j}$ is equivalent to sampling $\theta_{i,j}$ from $q(\theta_{i,j} | \phi_{i,j}, \alpha)$.  Now $\theta_{i,j}$ becomes a random variable parameterized by $\phi_{i,j}$.
\begin{align}
\theta_{i,j} & = \phi_{i,j} \xi_{i,j} = \phi_{i,j}(1+\sqrt{\alpha}\varepsilon_{i,j}) \\
\theta_{i,j} & \sim \mathcal{N}(\theta_{i,j} | \phi_{i,j}, \alpha \phi_{i,j}^2) \qquad  \varepsilon_{i,j} \sim \mathcal{N}(0,1) \notag
\label{eq:thetaDrop}
\end{align}

Variational Dropout uses $q(\theta  | \phi, \alpha)$ as an approximate posterior distribution for a model with a special prior on the weights. The variational parameters $\phi$ and $\alpha$ of the distribution $q(\theta  | \phi_{i,j}, \alpha)$ are tuned via stochastic variational inference as denoted in Section~\ref{sec:svi}. 
During dropout training, $\theta$ is adapted to maximize the expected log-likelihood~(\ref{eq:expectedLL}). 
For this to be consistent with the optimization of a variational lower bound, the prior on the weights $p(\theta)$ has to be such that $D_{KL} \big(q(\theta | \phi, \alpha) \parallel p(\theta)\big)$ does not depend on $\theta$.
The prior distribution that meets this requirement is chosen to be the scale invariant log-uniform prior~\cite{kingma2015variational}:
\begin{align}
p(\log\mid\theta_{i,j}\mid) = const  \Leftrightarrow  p(\log\mid\theta_{i,j}\mid) \propto {1} / {\mid\theta_{i,j}\mid}
\end{align}

Then, maximization of the variational lower bound~(\ref{eq:ELBO}) becomes equivalent to maximization of the expected log-likelihood with fixed parameter $\alpha$. 
It means that Gaussian Dropout training is exactly equivalent to Variational Dropout with fixed $\alpha$. However, Variational Dropout provides a way to train dropout rate $\alpha$ by optimizing the ELBO. Interestingly, dropout rate $\alpha$ now becomes a variational parameter and not a hyperparameter that allows us to train individual dropout rates $\alpha_{i,j}$ for each layer or even weight~\cite{kingma2015variational}.

\paragraph{\textbf{Additive Noise Reparameterization:}} 
Although the \emph{local reparameterization trick} reduces the variance of the variance of stochastic gradients, original multiplicative noise still yields large-variance gradients:
\begin{align*}
\theta_{i,j} = \phi_{i,j}(1+\sqrt{\alpha_{i,j}}\varepsilon_{i,j}), \quad \frac{\partial\theta_{i,j}}{\partial\phi_{i,j}} = 1+\sqrt{\alpha_{i,j}}\varepsilon_{i,j}, \quad \varepsilon_{i,j} \sim \mathcal{N}(0,1)  
\end{align*}
and very large values of $\alpha$ correspond to local optima from which it is hard to escape.
To avoid such local optima, Kingma \emph{et al.}~\cite{kingma2015variational} only considered the case of $\alpha \leq 1$, which corresponds to a binary dropout rate $p \leq 0.5$. 
However, the case of large $\alpha_{i,j}$ is very interesting (here we mean separate $\alpha_{i,j}$ per weight or neuron). Having high dropout rates ($\alpha_{i,j} \rightarrow +\infty$) corresponds to a binary dropout rate that approaches $p=1$. It effectively means that corresponding weight or a neuron is always ignored and can be removed from the model.
Molchanov \emph{et al.}~\cite{molchanov2017variational} introduced a trick that can train the model within the full range of $\alpha_{i,j} \in \left(0,+\infty\right]$ by reducing the variance of gradients even further. 
The idea is to replace the multiplicative noise term $1+\sqrt{\alpha_{i,j}} \varepsilon_{i,j}$ with an exactly equivalent additive noise term $\zeta_{i,j} \varepsilon_{i,j}$ where $\zeta_{i,j}^2 = \alpha_{i,j}\phi_{i,j}^2$ is treated as a new independent variable:
\begin{align}
\theta_{i,j} = \phi_{i,j}(1+\sqrt{\alpha_{i,j}}\varepsilon_{i,j}) = \phi_{i,j}+\zeta_{i,j}\varepsilon_{i,j} \ ,  \quad  \frac{\partial\theta_{i,j}}{\partial\phi_{i,j}} = 1 \ , \quad
 \varepsilon_{i,j} \sim \mathcal{N}(0,1) 
\label{eq:param}
\end{align}
After this trick, the ELBO is optimized w.r.t. $\theta$ and $\zeta$. However, $\alpha$ is still kept and used throughout the paper, since it has a nice interpretation as a dropout rate and can be obtained from $\alpha_{i,j} \phi_{i,j}^2 = \zeta_{i,j}^2$.
In addition to reducing the variance of gradients, this trick also helps us to propose a novel variational dropout algorithm that satisfies the differential privacy definition by adding independent Gaussian noise to the updates of each weight.

\section{Differentially Private Variational Dropout}
\label{sec:methods}
In this section, we describe our approach toward differentially private training of neural networks and introduce the proposed differentially private variational dropout (DPVD) algorithm. 
Algorithm~\ref{alg:PrivateDropout} outlines our basic method for training a model with parameters $\phi$ by minimizing the ELBO~(\ref{eq:ELBO}). 
At each iteration of the training scheme, DPVD takes a minibatch of data, samples the activations using the local reparameterization trick~\cite{kingma2015variational},  computes an estimate of the lower bound~(\ref{eq:ELBO}) and its gradient using parametrization trick~(\ref{eq:param}). 
This stochastic gradient is then used to update model parameters $\phi$ via some SGD-based optimization method by iteratively applying the following update equation at iteration $t$:
\begin{align}
\phi^{(t+1)} = \phi^{(t)} - \eta_t \tilde{g}
\label{eq:update2}
\end{align}
where $\eta_t$ is the learning rate and $\tilde{g}$ is the gradient of the lower bound $\tilde{g} = \big(\bigtriangledown_\theta \mathcal{L}(q_\phi)\big)$. 
Here, the goal is to estimate $\theta$ that equals to $\phi_{i,j} + \zeta_{i,j} \varepsilon_{i,j}$ as given in Section~\ref{sec:vd}. 
The weights $\theta$ are then computed with the following update rule:
\begin{align}
\theta_{i,j}^{(t+1)} = \phi_{i,j}^{(t)} + \zeta_{i,j} \varepsilon_{i,j} - \eta_t \tilde{g}
\label{eq:update2}
\end{align} 

The approach presented here regularizes the neural network by adding random noise $\zeta \varepsilon$. To protect the privacy of training data, the gradients need to be perturbed with Gaussian noise in each iteration; so we use the existing random noise in order to provide privacy.

At each step, we have chosen to perturb parameter updates with zero mean multivariate normal noise with covariance matrix $\sigma^2 C^2 \mathbb{I}$. 
The algorithm requires several parameters to determine the privacy budget. Sampling frequency $\nu$ for subsampling within the data set, a total number of iterations $T$ and clipping threshold $C$ are important design decisions.
Parameter $\sigma$ in noise level determines our total $\epsilon$ and depends on the total $\delta$ in privacy budget and clipping the gradients using the threshold $C$ will lead $L_2$ sensitivity of gradient sum to be $2C$. The amount of noise is chosen to be equal to the $\zeta \varepsilon$ in (\ref{eq:update2}).

\begin{algorithm}[h!]
\caption{Differentially Private Variational Dropout (DPVD)}
\begin{algorithmic}[1]
\STATE \textbf{Inputs}: Input data $\mathcal{D} = \{x_i, y_i\}_{i=1}^N$, number of data passes $T$, minibatch size $S$, learning rate $\eta_t$, noise scale $\sigma$, gradient norm bound $C$.
\STATE Initialize $\phi_0$ randomly
\FOR{ $t \leftarrow 1$ to $T$ }
\STATE Take a random sample $\mathcal{S}_t$ of size $S$ with sampling probability $\nu=S/N$ 
\STATE Compute gradient for each $(x_i,y_i) \in \mathcal{S}_t$: $\tilde{g}= \big(\bigtriangledown_\theta \mathcal{L}(q_\phi)\big)$ 
\STATE Clip gradient: $\bar{\tilde{g}} = \tilde{g}/\max\big(1, {\parallel\tilde{g}\parallel_2}/{C} \big)$ 
\STATE Compute noise: $\zeta \varepsilon \sim \mathcal{N}(0, 4 C^2 \sigma^2 \mathbb{I})$ 
\STATE Update parameter: $\theta^{(t+1)} \leftarrow \phi^{(t)} + \zeta \varepsilon - \eta_t \bar{\tilde{g}}$
\ENDFOR
\STATE \textbf{Output}: $\theta^{(T)}$.
\end{algorithmic}
\label{alg:PrivateDropout}
\end{algorithm}

In this work, we first calculate the per-iteration privacy budget using the key properties of advanced composition theorem (Theorem 3.20 of \cite{Dwork2014}) and this method is called \emph{DPVD-AC} in the experiments. Then, we use a relaxed notion of differential privacy, called zCDP~\cite{bun2016concentrated} that bounds the moments of the privacy loss random variable and call this method \emph{DPVD-zCDP} in the experiments. The moments bound yields a tighter tail bound, and consequently, it allows for a higher per-iteration budget than standard DP-methods for a given total privacy budget. The description of how we chose the privacy design parameters and calculate the privacy budget is given in Section 4.3~\cite{ermis2017differentially}.

\section{Experiments and Results}
\label{sec:exp_res}
We evaluate our approach on two standard benchmark datasets. MNIST dataset~\cite{Lecun98} contains 70K $28 \times 28$ hand-written digits (60K for training and 10K for testing). DIGITS~\cite{bache2013uci} dataset consists of 1797 $8 \times 8$ grayscale images (1439 for training and 360 for testing) of handwritten digits. 
We use a simple feed-forward neural network with ReLU units and softmax of 10 classes for both datasets.
All experiments are implemented in Theano~\cite{theano2016}. 

\paragraph{\textbf{Baseline: }} Our baseline models use a single hidden layer with 1000 hidden units. For MNIST dataset, we use the minibatch of size 600 and reach an  accuracy of $97.80\%$ in about 200 epochs. For DIGITS dataset, we use the minibatch of size 100 and reach an accuracy of $95.35\%$ in about 100 epochs. All of the results in this section are the average of 10 runs.
\begin{figure}[h!]
\begin{minipage}[b]{0.50\textwidth}
\centering
\subfigure[DPVD-AC]{\includegraphics[scale=0.25]{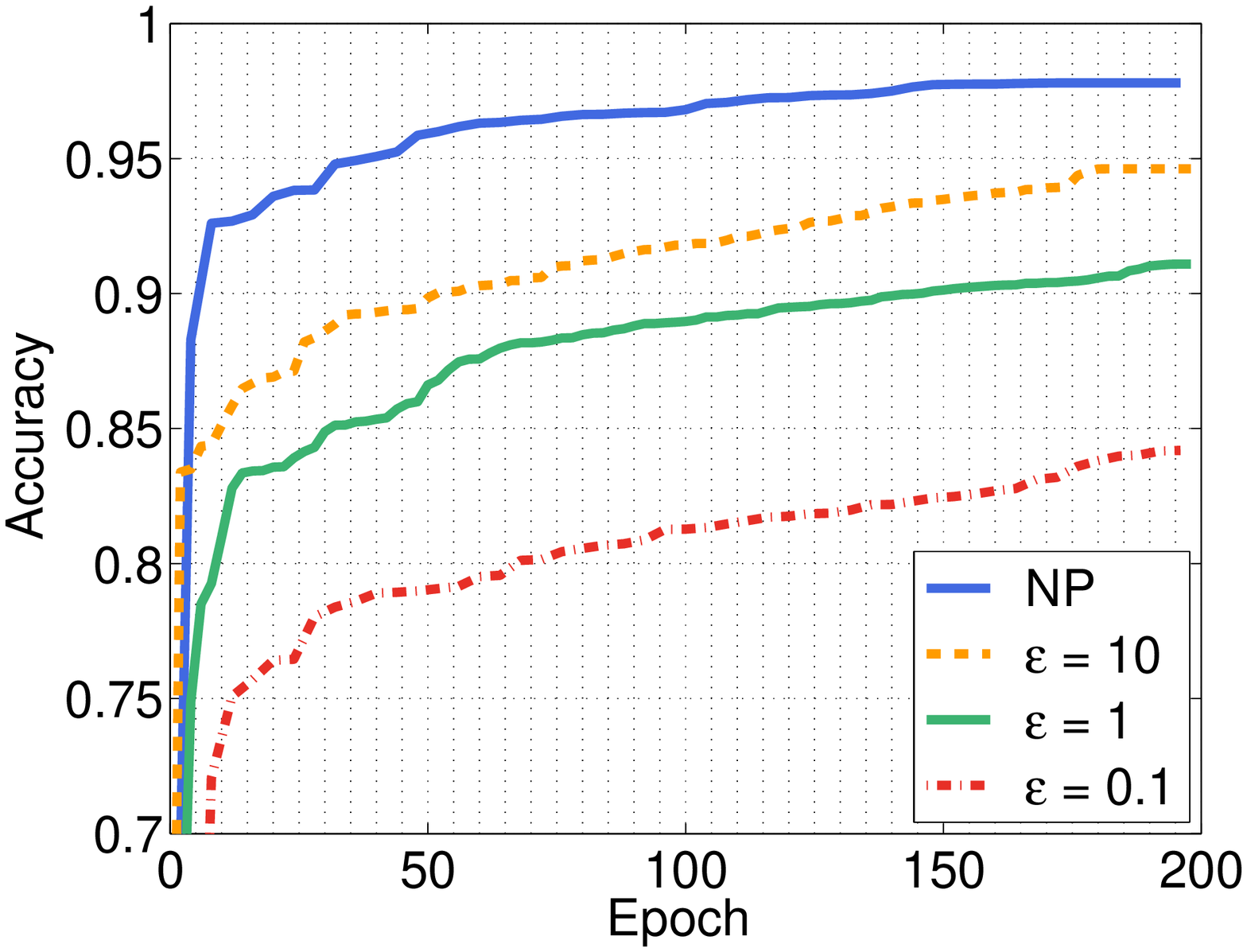} \label{subfig:DPmnist}}
\end{minipage}   
\begin{minipage}[b]{0.50\textwidth} 
\centering
\subfigure[DPVD-zCDP]{\includegraphics[scale=0.25]{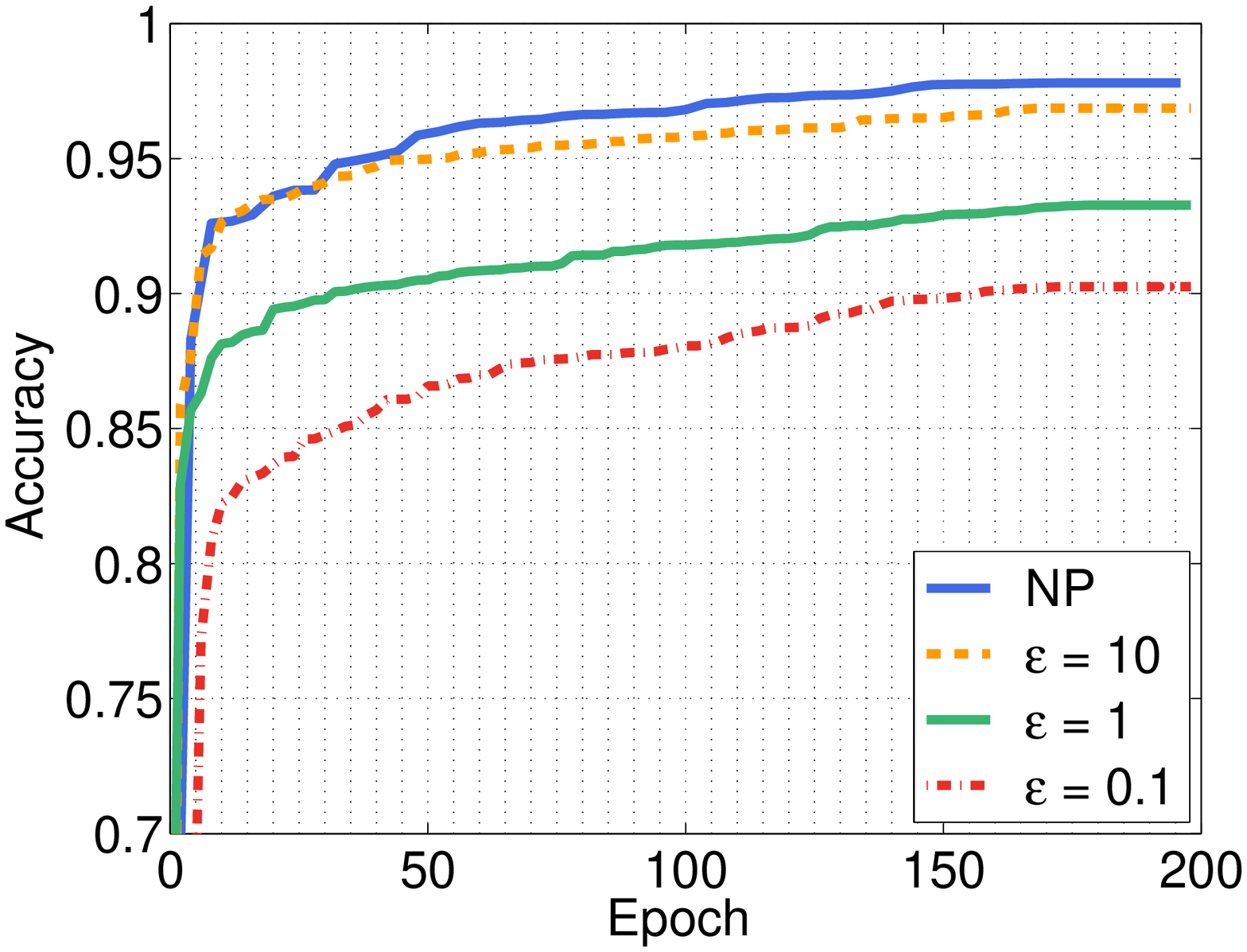} \label{subfig:zcDPmnist}}
\end{minipage}
\caption{Comparison of the test accuracies on MNIST for $\epsilon=\lbrace 10, 1, 0.1\rbrace$ and the NP case.}
\label{fig:Mnist}
\end{figure}

\paragraph{\textbf{Differentially private models: }} For the differentially private version, we experiment with the same architecture. To limit sensitivity, we clip the gradient norm of each layer at $C=2$. We report results for three different noise levels where $\epsilon = \lbrace 10, 1, 0.1\rbrace$. 
For any fixed $\epsilon$, $\delta$ is varied between $10^{-2}$ and $10^{-5}$. There is a slight difference with different $\delta$ values (less than $10^{-3}$), but still we choose the best performing $\delta=10^{-5}$ for both datasets. We set the initial learning rate $\eta_0=0.1$ for MNIST and $\eta_0=0.05$ for DIGITS and update it per round as $\eta_t = \eta_0/t^\gamma$. We fixed the decay rate $\gamma$ to 1 for both of the datasets. 
\begin{figure}[h!]
\begin{minipage}[b]{0.50\textwidth}
\centering
\subfigure[DPVD-AC]{\includegraphics[scale=0.25]{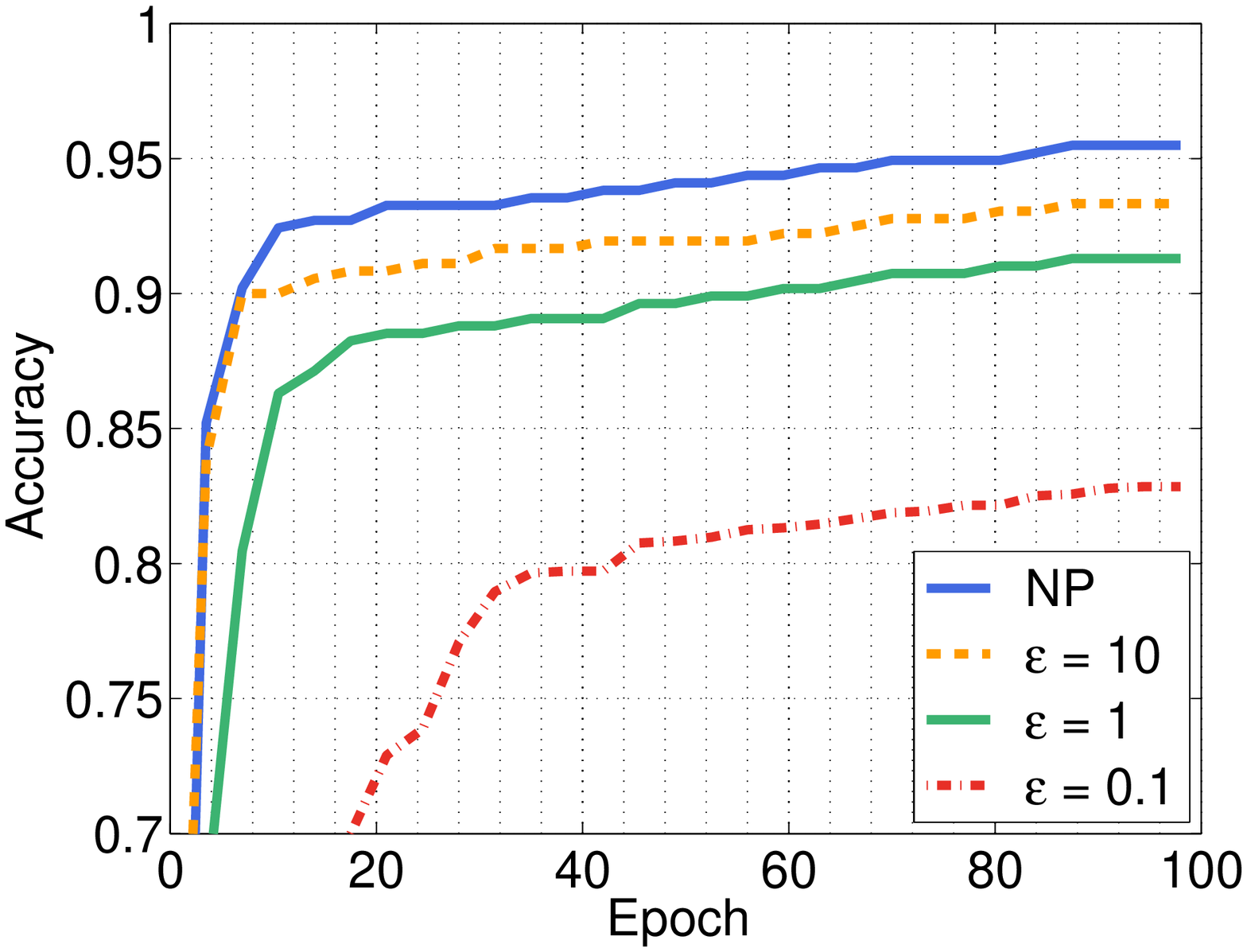} \label{subfig:DPdigits}}
\end{minipage}   
\begin{minipage}[b]{0.50\textwidth} 
\centering
\subfigure[DPVD-zCDP]{\includegraphics[scale=0.25]{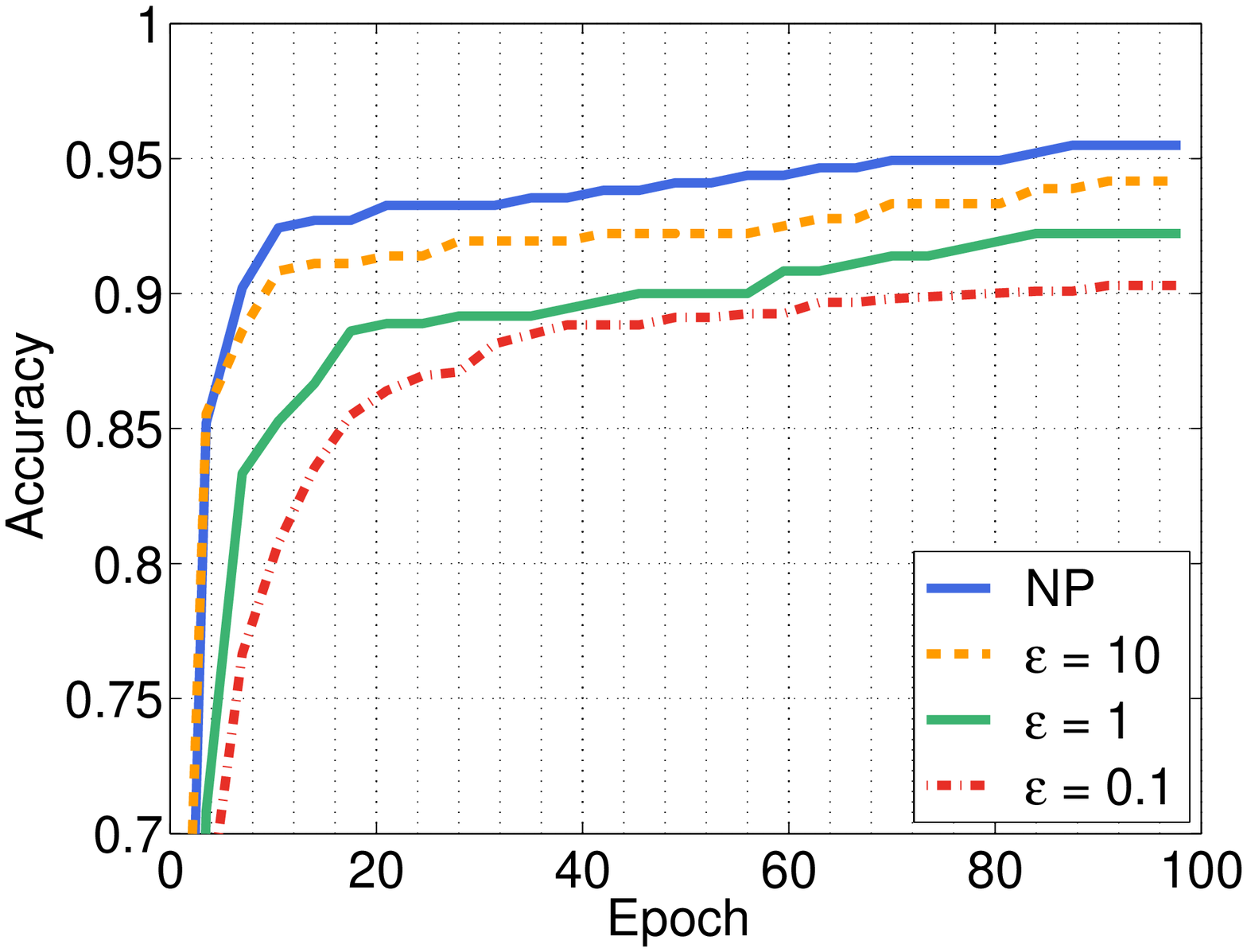} \label{subfig:zcDPdigits}}
\end{minipage}
\caption{Comparison of the test accuracies on DIGITS for $\epsilon = \lbrace 10, 1, 0.1\rbrace$ and the NP case.}
\label{fig:Digits}
\vspace*{-5mm}
\end{figure}

In the first set of experiments, we investigate \emph{the influence of privacy loss on accuracy}. The mini-batch sizes were set to $S=600$ and $S=100$ for MNIST and DIGITS respectively. We ran the algorithm for 200 passes for MNIST and 100 passes for DIGITS. Figure~\ref{subfig:DPmnist},~\ref{subfig:DPdigits} report the performance of the DPVD-AC method and Figure~\ref{subfig:zcDPmnist},~\ref{subfig:zcDPdigits} report the performance of the DPVD-zCDP method for different noise levels. 
These results justify the theoretical claims that lower prediction accuracy is obtained when the privacy protection is increased by decreasing $\epsilon$. 
One more deduction from these results is that variational dropout with the zCDP composition can reach an accuracy very close to the non-private level especially under reasonably strong privacy guarantees (when $\epsilon > 0.1$).
\begin{figure}[h!]
\begin{minipage}[b]{0.50\textwidth} 
\centering
\subfigure[MNIST]{\includegraphics[scale=0.25]{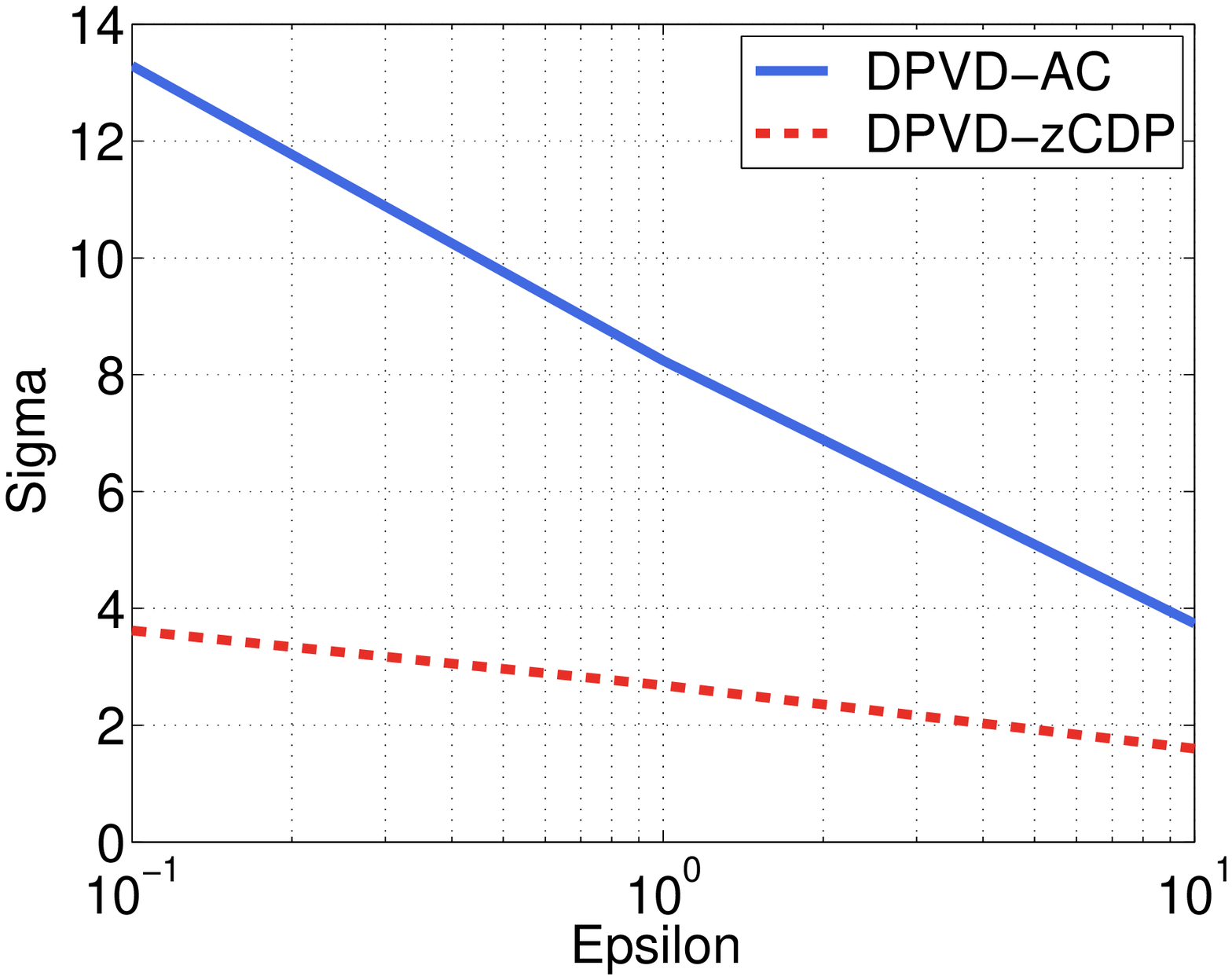} \label{subfig:compSigmaM}}
\end{minipage}
\begin{minipage}[b]{0.50\textwidth} 
\centering
\subfigure[DIGITS]{\includegraphics[scale=0.25]{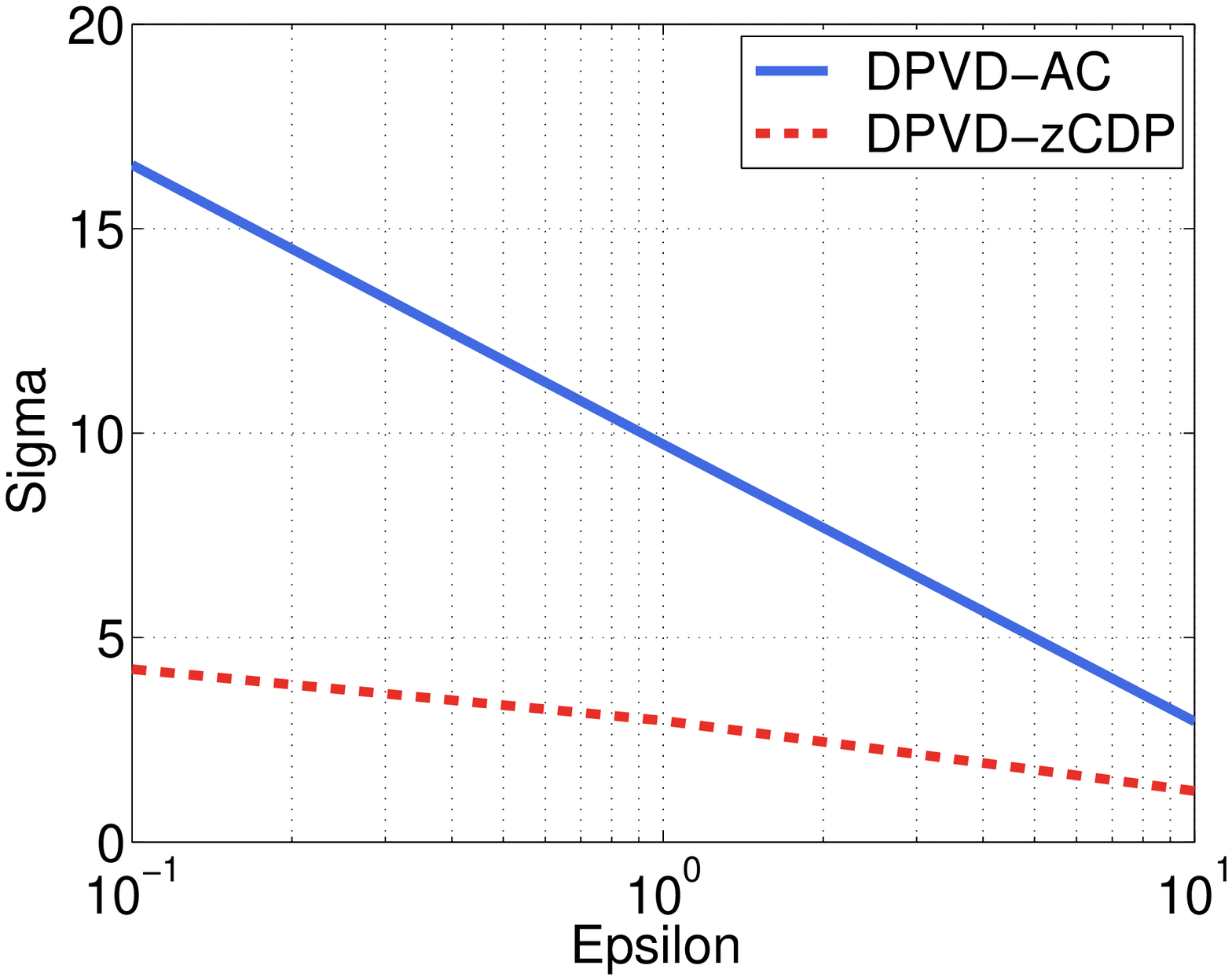} \label{subfig:compSigmaD}}
\end{minipage}
\caption{The $\sigma$ value as a function of $\epsilon$.}
\label{fig:compareSigma}
\end{figure}

Then, we compare the classification accuracy of models learned using two variants of the variational dropout algorithm: DPVD-AC and DPVD-zCDP. As mentioned in Section~\ref{sec:dp_budget}, zCDP composition provides a tighter bound on the privacy loss compared to the advanced composition theorem. 
Here we compare them using some concrete values. The noise level can be computed from the overall privacy loss $\epsilon$, the sampling ratio of each minibatch $\nu$ = $S/N$ and the number of epochs $E$ (so the number of iterations is $T$ = $E/\nu$). For our MNIST and DIGITS experiments, we set $\nu$ = 0.01, $E$ = 200 and $\nu$ = 0.05, $E$ = 100, respectively. Then, we compute the value of $\sigma$. 
For example, when $\epsilon$ = 1, the $\sigma$ values are 8.24 for DPVD-AC and 2.68 for DPVD-zCDP on MNIST, and the $\sigma$ values are 9.73 for DPVD-AC and 2.97 for DPVD-zCDP on DIGITS. We can see from Figure~\ref{fig:compareSigma} that we get a much lower noise by using the zCDP for a fixed the privacy loss $\epsilon$. 
\begin{figure}[h!]
\begin{minipage}[b]{0.50\textwidth} 
\centering
\subfigure[MNIST]{\includegraphics[scale=0.25]{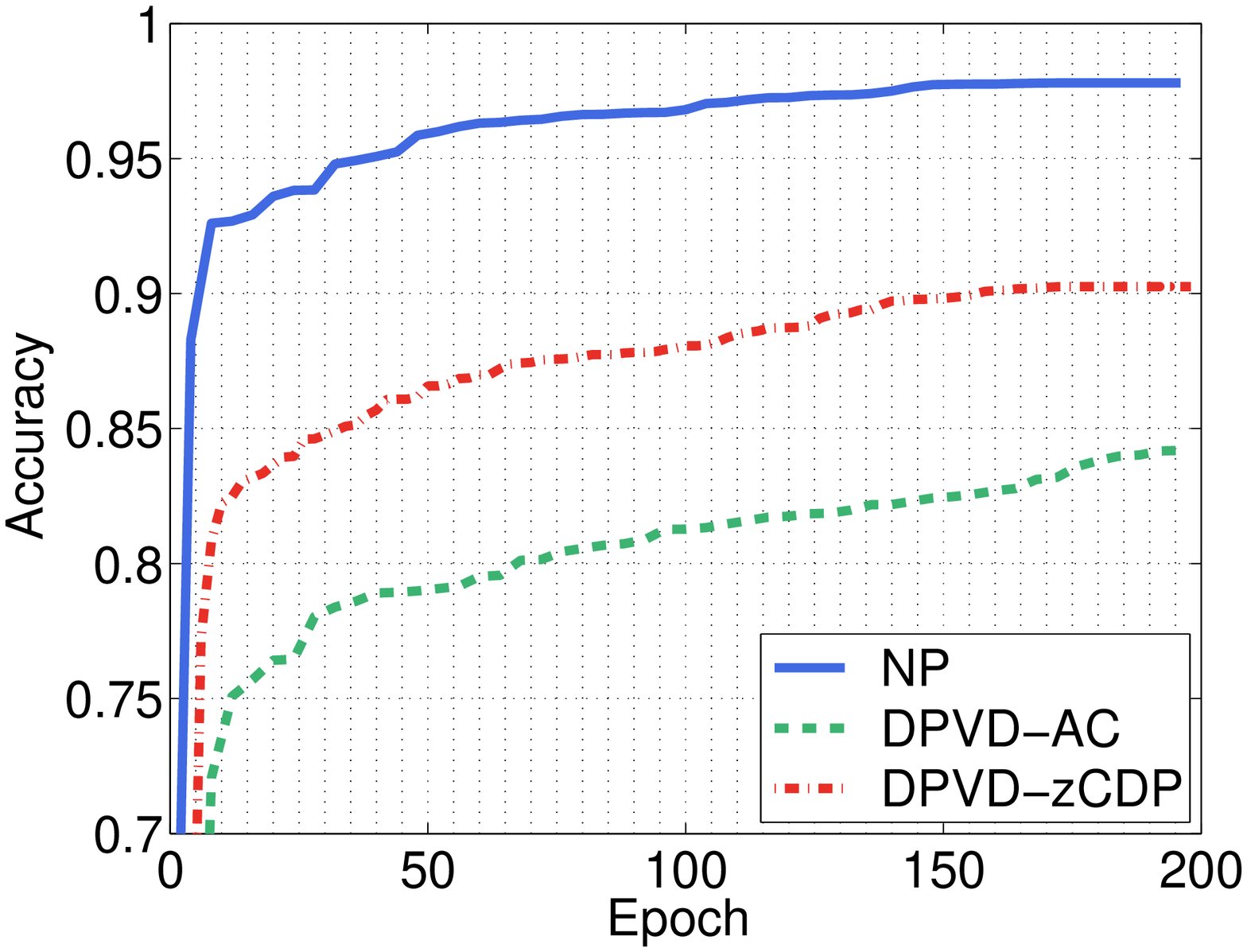} \label{subfig:comp3mnist}}
\end{minipage}
\begin{minipage}[b]{0.50\textwidth} 
\centering
\subfigure[DIGITS]{\includegraphics[scale=0.25]{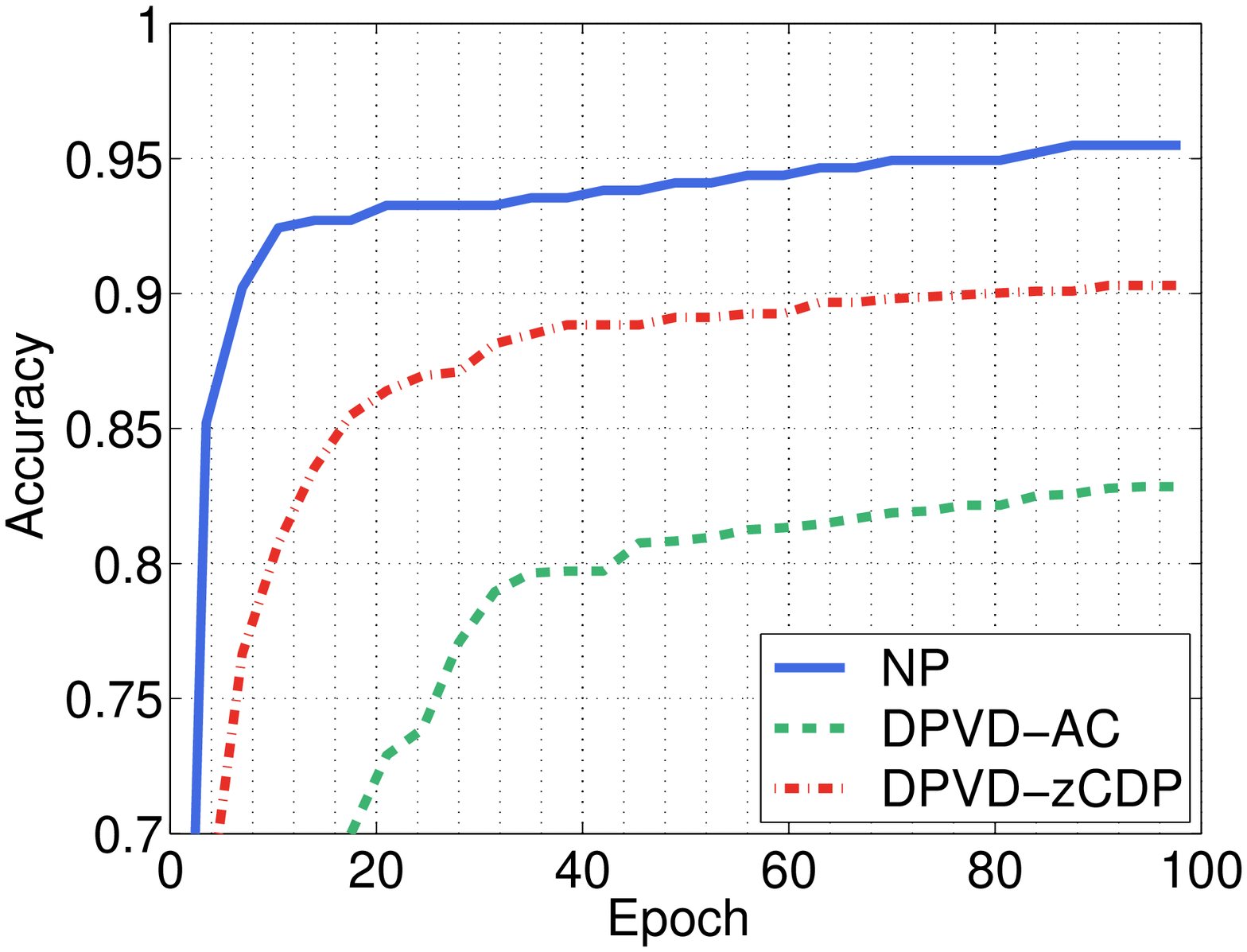} \label{subfig:comp3digits}}
\end{minipage}
\caption{Test accuracy results of DPVD-AC and DPVD-zCDP for $\epsilon$ = 0.1. }
\label{fig:compare}
\end{figure}
Therefore, for our neural network models with a total privacy budget fixed to $\epsilon$, the amount of noise added is smaller for zCDP, and the test accuracy is higher. Figure~\ref{fig:compare} shows the comparison results of DPVD-AC and DPVD-zCDP methods when $\epsilon$ = 0.1 with non-private model. Both results clearly show that using the zCDP composition further helps in obtaining even more accurate results at a comparable level of privacy.

We compare our methods to the most related algorithm proposed by Abadi \emph{et al.}~\cite{Abadi2016} and the case when no dropout is used. For the algorithm with no dropout, we use SVI to update the weights of the neural network. We ran all the methods on MNIST and DIGITS with varying $\epsilon$. Table~\ref{table:Comparison} reports the test accuracies of all methods. The previous experiments have already demonstrated that DPVD-zCDP significantly improves the classification accuracy. These results also support it and show that \emph{dropout} improves the prediction accuracy on differentially private neural networks especially when the privacy budget is low.
\renewcommand{\arraystretch}{1.2}
\begin{table}[ht!]
\caption{Comparison of the methods for $\epsilon = \lbrace 10, 1, 0.1\rbrace$. Bold values indicate the best results.}
\begin{center}
\begin{tabular}{|  L{31mm}  | C{13mm}  | C{13mm}  | C{13mm} ||  C{13mm}  |  C{13mm}  |  C{13mm} |}
\hline     &   \multicolumn{3}{c ||}{MNIST }   &  \multicolumn{3}{c |}{DIGITS }    \\  \cline{2-7}
              &  $\epsilon$ = 10   &   $\epsilon$ = 1     &   $\epsilon$ = 0.1    &   $\epsilon$ = 10   &   $\epsilon$ = 1    &   $\epsilon$ = 0.1  \\  \hline
DPVD-AC                                               &  0.9462   &    0.9102                 &  0.8419                  &  0.9361    &                0.9139     &   0.8217    \\  \cline{1-7}
DPVD-zCDP                                           &  0.9687   &    \textbf{0.9327}   &  \textbf{0.9026}    &  0.9417    &   \textbf{0.9278}   &   \textbf{0.9038}   \\  \cline{1-7}
SVI-zCDP (no dropout)                          &  0.9518   &    0.9126                 &  0.8791                  &  0.9375     &                0.9107    &   0.8712    \\  \cline{1-7}
Abadi \emph{et al.}~\cite{Abadi2016} &  \textbf{0.9701}  &    0.9305    &  0.8875                  &  \textbf{0.9450}    &   0.9265    &   0.8943   \\ \hline
\end{tabular}
\label{table:Comparison}
\end{center}
\vspace*{-8mm}
\end{table}

\begin{figure}[h!]
\begin{minipage}[b]{0.50\textwidth}
\centering
\subfigure[Number of hidden units]{\includegraphics[scale=0.25]{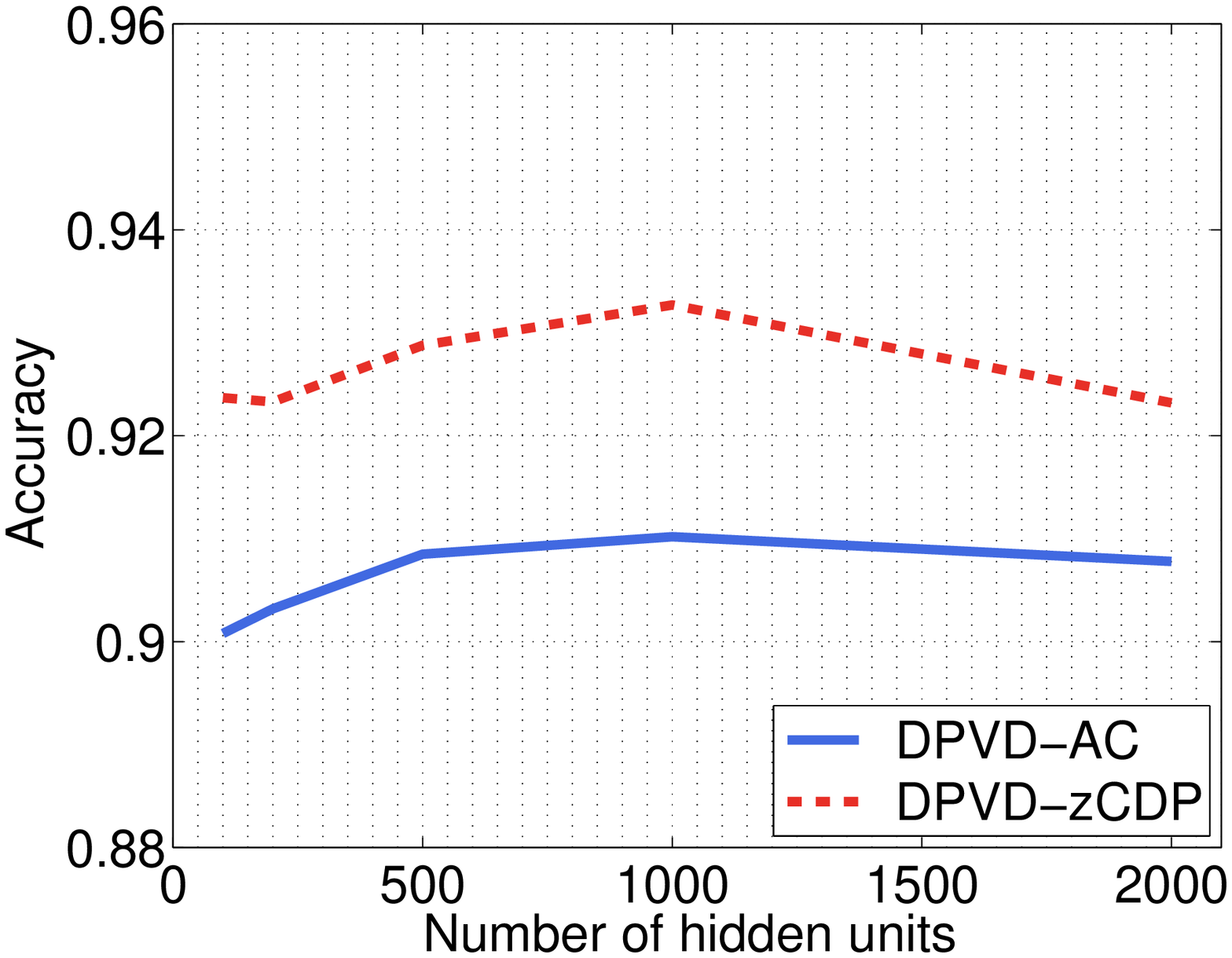} \label{subfig:compHiddenU}}
\end{minipage}   
\begin{minipage}[b]{0.50\textwidth}
\centering
\subfigure[Number of epochs]{\includegraphics[scale=0.25]{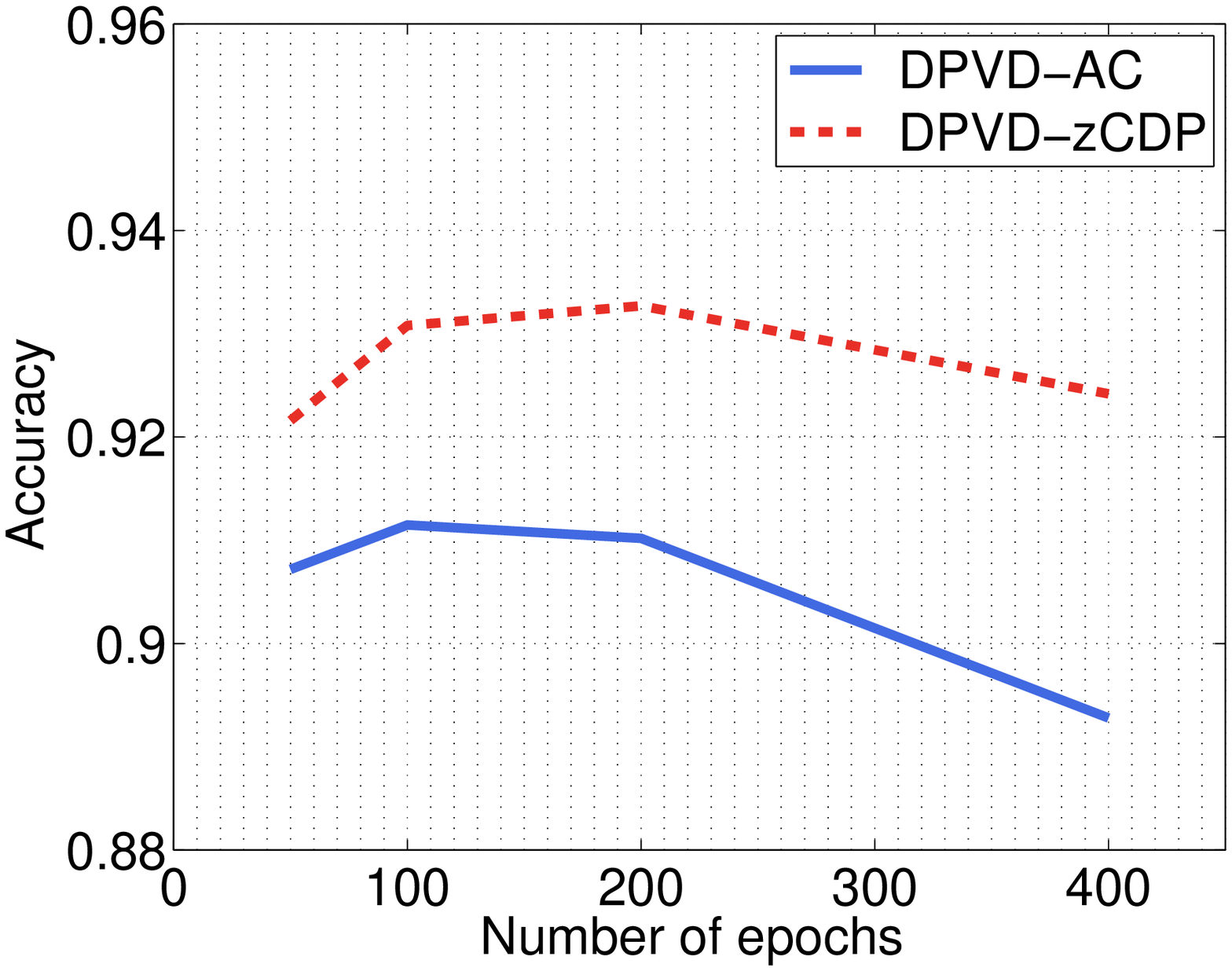} \label{subfig:compEpoch}}
\end{minipage}
\\   
\begin{minipage}[b]{0.50\textwidth} 
\centering
\subfigure[Gradient clipping norm]{\includegraphics[scale=0.25]{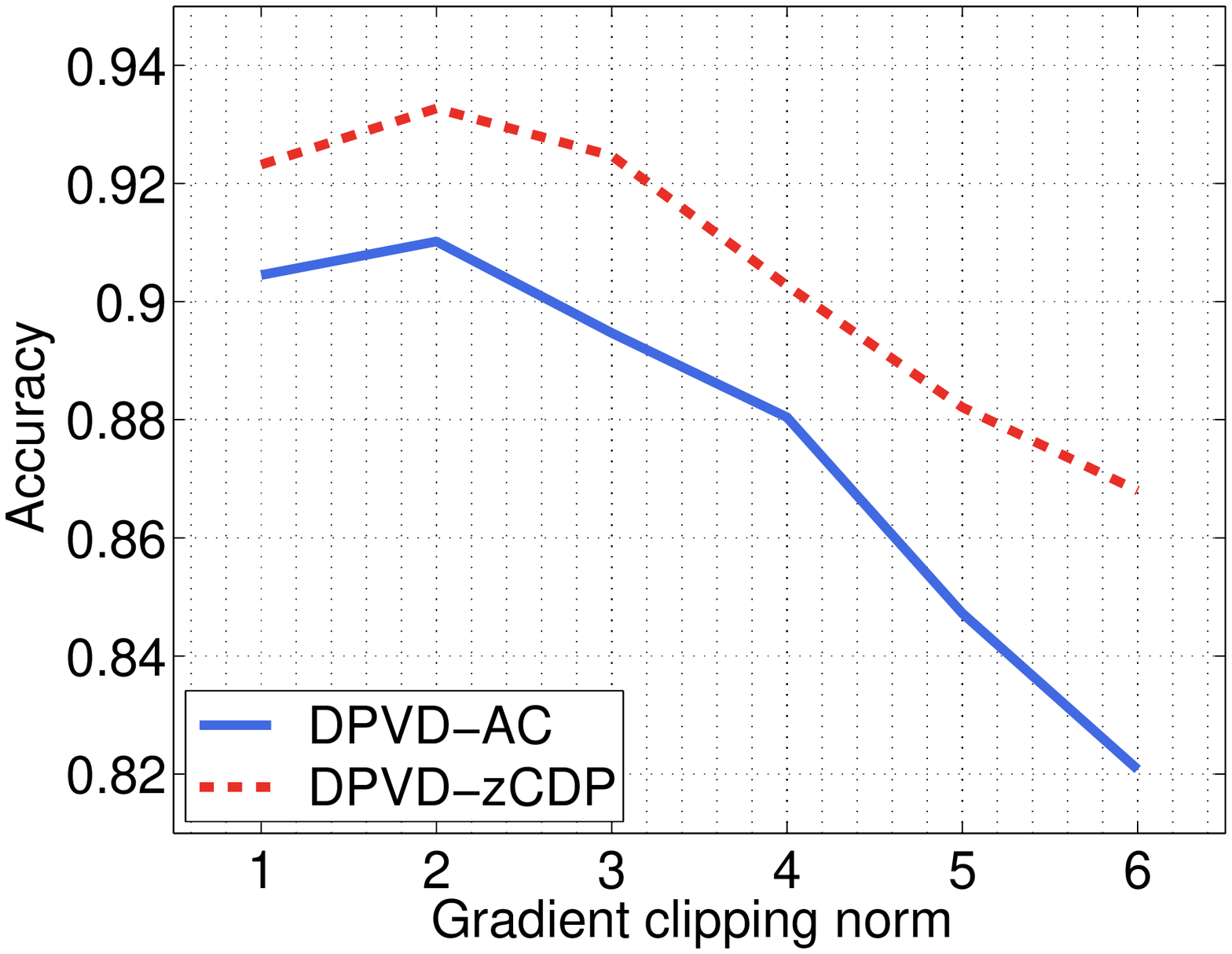}  \label{subfig:compGradClip}}
\end{minipage}
\begin{minipage}[b]{0.50\textwidth} 
\centering
\subfigure[Minibatch size]{\includegraphics[scale=0.25]{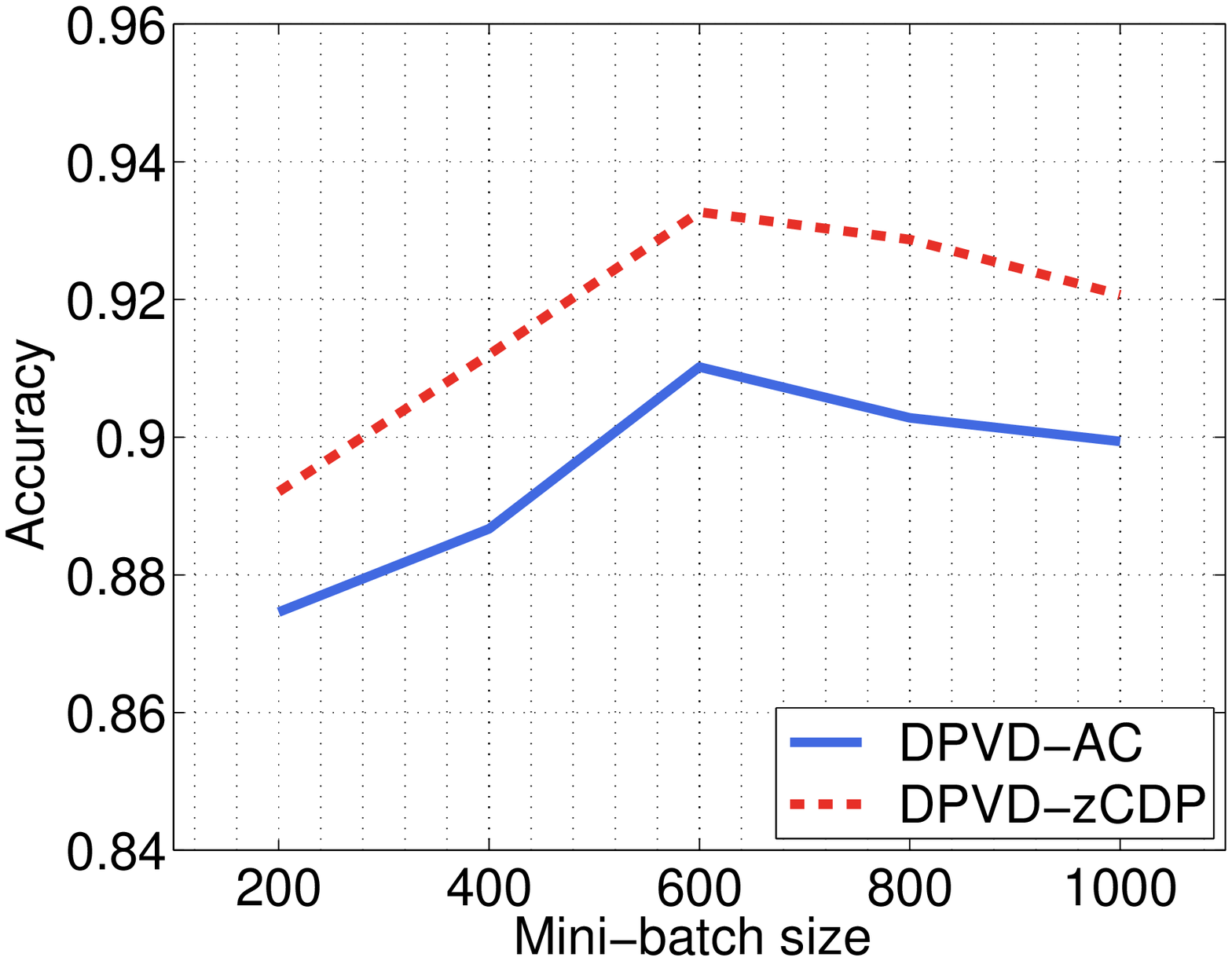} \label{subfig:compMbSize}}
\end{minipage}
\caption{Effect of the model parameters on MNIST dataset.}
\label{fig:MnistPar}
\end{figure}

\paragraph{\textbf{Effect of the parameters: }}
The classification accuracy in neural networks depends on a number of parameters that must be carefully tuned to optimize the performance. For our differentially private models, these factors include the number of hidden units, the number of iterations, the gradient clipping threshold, the minibatch size and the noise level. In the previous section, we compared the effect of different noise levels to the classification accuracy. 
Here, we demonstrate the effects of the remaining parameters. We control a parameter individually by fixing the rest as constant. For MNIST experiments, we set the parameters as follows: 1000 hidden units, minibatch size of 600, gradient norm bound of 2, initial learning rate of 0.1, 200 epochs and the privacy budget $\epsilon$ to 1. 
The results are presented in Figure~\ref{fig:MnistPar}. 

In standard, non-private neural networks, using more hidden units is often preferable and increases the prediction accuracy of the trained model. For differentially private training, using more hidden units leads more noise added at each update due to the increase in the sensitivity of the gradient. However, increasing the number of hidden units does not always decrease accuracy since larger networks are more tolerant to noise. Figure~\ref{subfig:compHiddenU} shows that accuracy is very close for a hidden unit number in the range of $\left[500, 2000\right]$ and peaks at 1000. 

The number of epochs $E$ (so the number of iterations is $T$ = $E/\nu$) needs to be sufficient but not too large. The privacy cost in zCDP increases in proportion to $\sqrt{T}$ and it is more tolerable than DP. We tried several values in the range of $\left[50, 400\right]$ and observed that we obtained the best results when $E$ is between 100 and 200 for MNIST.

Tuning the gradient clipping threshold depends on the details of the model. If the threshold is too small, the clipped gradient may point in a very different direction from the true gradient. Besides, when we increase the threshold, we add a large amount of noise to the gradients. Abadi \emph{et al.}~\cite{Abadi2016} proposed that a good way to choose a value for $C$ is taking the median of the norms of the unclipped gradients. In our experiments, we tried $C$ values in the range of $\left[1, 6\right]$. Figure~\ref{subfig:compGradClip} shows that our model is tolerable to the noise up to $C$ = 4, then the accuracy decreases marginally.

Finally, we monitor the effect of the minibatch size. In DP settings, choosing smaller minibatch size leads running more epochs, however, the added noise has a smaller relative effect for a larger minibatch. Figure~\ref{subfig:compMbSize} shows that relatively larger minibatch sizes give better results. Empirically, we obtain the best accuracy when $S$ is around 600 (so the sampling frequency $\nu = S/N = 0.01$). Due to space limitations, we only present the results on MNIST data, the results on DIGITS data have very similar behavior.

\section{Conclusion}
\label{sec:conc}
We introduced differentially private variational dropout method that outputs privatized results with accuracy close to the non-private inference results, especially under reasonably strong privacy guarantees. To make effective use of the privacy budget over multiple iterations, we proposed to calculate the cumulative privacy cost by using zCDP. Then, we showed how to perform variational dropout method in private settings. We illustrated the effectiveness of our algorithm on several benchmark datasets. One natural next step is to extend the approach to distributed training of neural networks. The algorithm proposed in the paper are generic and it can be applied to any neural network model. We left its application to other variants of neural networks such as convolutional and recurrent neural networks to future work.


\bibliographystyle{splncs}
\bibliography{paper}

\end{document}